\documentclass[runningheads,a4paper]{llncs}[2014/03/31 v2.19]

\usepackage{cmap}

\iffalse 
\usepackage[%
rm={oldstyle=false,proportional=true},%
sf={oldstyle=false,proportional=true},%
tt={oldstyle=false,proportional=true,variable=true},%
qt=false%
]{cfr-lm}
\else
\usepackage{newtxtext}
\usepackage{newtxmath}
\usepackage[zerostyle=b,scaled=.9]{newtxtt}
\fi

\usepackage[math]{blindtext}
\usepackage{mwe}

\usepackage[T1]{fontenc}
\usepackage[utf8]{inputenc} 

\usepackage{graphicx}

\usepackage[ngerman,english]{babel}
\addto\extrasenglish{\languageshorthands{ngerman}\useshorthands{"}}

\usepackage{upquote}

\usepackage{booktabs}

\usepackage{paralist}

\usepackage{csquotes}

\RequirePackage{iftex}
\ifPDFTeX
\else
\fi%

\usepackage{url}
\makeatletter
\g@addto@macro{\UrlBreaks}{\UrlOrds}
\makeatother

\usepackage[usenames,dvipsnames,svgnames,table]{xcolor}

\usepackage{listings}
\renewcommand{\lstlistingname}{List.}
\usepackage{chngcntr}
\AtBeginDocument{\counterwithout{lstlisting}{section}}

\AtBeginEnvironment{listing}{\setcounter{listing}{\value{lstlisting}}}
\AtEndEnvironment{listing}{\stepcounter{lstlisting}}

\usepackage{pdfcomment}
\usepackage[%
square,        
comma,         
numbers,       
sort&compress, 
]{natbib}

\usepackage[capitalise,nameinlink]{cleveref}
\crefname{section}{Sect.}{Sect.}
\Crefname{section}{Section}{Sections}
\crefname{listing}{\lstlistingname}{\lstlistingname}
\Crefname{listing}{Listing}{Listings}

\usepackage{ltxcmds}
\makeatletter
\newcommand{\IfPackageLoaded}[2]{\ltx@ifpackageloaded{#1}{#2}{}}
\makeatother

\IfPackageLoaded{minted}{
	\setminted{numbersep=5pt, xleftmargin=12pt}
	
	\usemintedstyle{friendly}
	
	\usepackage[labelfont=bf,font=small,skip=4pt]{caption}
	\SetupFloatingEnvironment{listing}{name=List.,within=none}
}{
}
\ifcsmacro{listing}{}{
	\newenvironment{listing}[1][htbp!]{\begin{figure}[#1]}{\end{figure}}
	\newcounter{listing}
}
\ifcsmacro{minted}{}{%
	
}

\usepackage{xspace}

\DeclareFontFamily{U}{MnSymbolC}{}
\DeclareSymbolFont{MnSyC}{U}{MnSymbolC}{m}{n}
\DeclareFontShape{U}{MnSymbolC}{m}{n}{
	<-6>    MnSymbolC5
	<6-7>   MnSymbolC6
	<7-8>   MnSymbolC7
	<8-9>   MnSymbolC8
	<9-10>  MnSymbolC9
	<10-12> MnSymbolC10
	<12->   MnSymbolC12%
}{}
\DeclareMathSymbol{\powerset}{\mathord}{MnSyC}{180}

\usepackage[caption=false,font=normalsize,labelfont=sf,textfont=sf]{subfig}
\usepackage{capt-of}

\usepackage{siunitx}

\usepackage{amssymb}

\begin{document}
	\title{Exploring Adversarial Examples
	}
\subtitle{Patterns of One-Pixel Attacks}





 \author{%
     David Kügler\inst{1} \and
     Alexander Distergoft\inst{1} \and
     Arjan Kuijper\inst{2} \and
     Anirban Mukhopadhyay\inst{1}
 }
%
 \authorrunning{David Kügler, Alexander Distergoft, Arjan Kuijper, Anirban Mukhopadhyay}
  \institute{      
      $^1$ Interactive Graphics Systems Group, Technische Universität Darmstadt, Darmstadt, Germany\\
      $^2$ Fraunhofer IGD, Darmstadt, Germany\\
}

\maketitle

\begin{abstract}
	Failure cases of black-box deep learning, e.g. adversarial examples, might have severe consequences in healthcare. 
	Yet such failures are mostly studied in the context of real-world images with calibrated attacks. 
	To demystify the adversarial examples, rigorous studies need to be designed. 
	Unfortunately, complexity of the medical images hinders such study design directly from the medical images. 
	We hypothesize that adversarial examples might result from the incorrect mapping of image space to the low dimensional generation manifold by deep networks. 
	To test the hypothesis, we simplify a complex medical problem namely pose estimation of surgical tools into its barest form. 
	An analytical decision boundary and exhaustive search of the one-pixel attack across multiple image dimensions let us localize the regions of frequent successful one-pixel attacks at the image space.
\end{abstract}

\begin{keywords}
	CNN, Adversarial Examples, One-pixel Attack, Deep Learning Fails
\end{keywords}
\section{Introduction}
End-to-end Deep Learning pipelines (image in, classification out) have achieved significant success in Medical Image Computing (MIC) across multiple scenarios, even stretching to Computer-Aided Interventions (CAI) \cite{i3PosNet.2018}.
This success in comparison to traditional methods (including based on learning) has hurried an AI-summer in Healthcare seen in the prevalence of Deep Learning-publications in the MICCAI community. 
Political authorities have recognized this shift towards Deep Learning-based methods and are taking action.
In the United States, the FDA has embraced this change by approving AI devices for diabetic retinopathy detection \cite{Reuters.20180411} and is currently in the discussion towards easing the approval process for AI-based medical software \cite{FDAclassifyAI}. 
The European Union, on the other hand, has introduced the General Data Protection Regulation, 
which necessitates the right to explanation of any decisions taken by a computerized system. 

Since researchers struggle to explain decisions by Deep Learning models, the underlying function is yet a Black Box in practice.
Its analysis is hindered by the difficulty of deriving and understanding the decision boundary.
In fact, recent studies \cite{Rauber.20180320,Eykholt.20180410,Nguyen.20150402,Goodfellow.20150320,Szegedy.20140219} have shown that these Deep Learning models are vulnerable to adversarial examples -- 
these are images which cause incorrect classifications despite either models predicting with high certainty or being clear classifications to humans. 
Adversarial examples are not understood as a consequence of the black-box-characteristic. 
They can even be as simple as only changing a single pixel leading to different classification results (one-pixel-attacks).
For medical applications, the exploitation of this vulnerability is thoroughly analyzed by Finlayson \emph{et al.} \cite{Finlayson.20180521}.
However, this vulnerability is largely ignored across evaluations of Deep Learning in MIC and CAI.

Though impressive, the general example attacks shown by Finlayson \emph{et al.} \cite{Finlayson.20180521} address the traditional image-in-diagnosis-out setting. 
However, real medical images and annotations are complex further obscuring the situation and adding to the mystery
-- for example complex image structure, confounding situations (device vendors, acquisition parameters), non-conformity between radiologists, multi-class and multi-label decisions.
Without disentangling these factors it is impossible to understand adversarial examples, which stem from Deep Learning only.
Simplifications of MIC scenarios are needed to draw systematic conclusions regarding adversarial examples.
For example: 
to consider segmentation masks rather than the real-world images (binary instead of continuous pixel values) or to define decision boundaries in a closed form (which is not available in MIC).


In this paper, we provide the first systematic analysis of one-pixel-attacks on convolutional neural networks (CNNs) in a simplified CAI application and provide a first intuition of patterns.
With inherent limits on the knowledge and processable size of both the likely image space and a complete description of the decision boundary, it will be impossible to analyze adversarial attacks due to the complexity of CNNs -- even here simplifications are needed.
To break the problem down to its core, we 
a) simplify the range of images (image space), 
b) train multiple classifiers and
c) search exhaustively for one-pixel adversarial examples.
We consider the problem of instrument pose estimation studied by Kügler \emph{et al.} \cite{i3PosNet.2018}, who have ignored adversarial attacks, where the orientation of instruments is to be determined.
To gain control over the image and annotation complexity, we define a continuous generation manifold with a perfectly defined binary decision boundary. 
From individual manifold coordinates, we generate binary images at various dimensions with the instrument being simplified to a line for different levels of discretization.
We define all images that can be generated through this pipeline as ``possible images'' and train multiple simple classifiers based on convolutional neural networks with ``ALL'' these uniquely possible images. 
Finally, we exhaustively search the space of all single pixel-flip adversarial candidates, identify successful attacks and localize the regions of frequent successful one-pixel-attacks. 
The most surprisingly, the overwhelming majority of attacks are localized at a distance of the instrument, which implies the one-pixel-flip \emph{did not change the information of the image}.
	
\section{Related Work}
Goodfellow \emph{et al.}\cite{Goodfellow.20150320} demonstrate that standard image models exhibit a strange phenomenom: most randomly chosen images from a data distribution are correctly classified and yet are close to visually similar images that are incorrectly classified.
By adding some certain kind of pertubation to an image this behaviour can be reproduced on most CNNs. 
A hypothesis on that behaviour is that neural network classifiers are too linear in various regions of the input space.

Su \emph{et al.} \cite{Su.20180222} specialize on so called 'one pixel attacks'. They show that even by adding a pertubation with the size a single pixel to an image, the output of deep learning networks can easily be altered.

Gilmer \emph{et al.} \cite{Gilmer.20180111} try to get an insight on adversarial examples by training different neural networks on a synthetic dataset of two concentric spheres with different radii.
The idea is to classify whether a point belongs to one or the other sphere.
Even though the data manifold as well as the theoretical max margin boundary are clearly defined and enough input is provided for networks to train on, adversarials can still be found near correctly classified points.

\section {Methods}
We generate a custom dataset to study adversarial examples for image classification.
Using a exhaustive search, all one-pixel-flip candidate images are tested for misclassification.

Analyzing the space $\mathbb{I}$ of images, we differentiate between images belonging to an application ($\mathbb{I^*_f} \subset \mathbb{I}$, often termed ``natural images'') and images holding no information on the application.
By introducing a generation manifold $\mathbb{M}$ -- a higher-level parameter space describing all images possible for our application -- we are able to describe all features of the image relevant to our application.

\begin{figure}[t]
	\centering
	\includegraphics[width=0.8\textwidth]{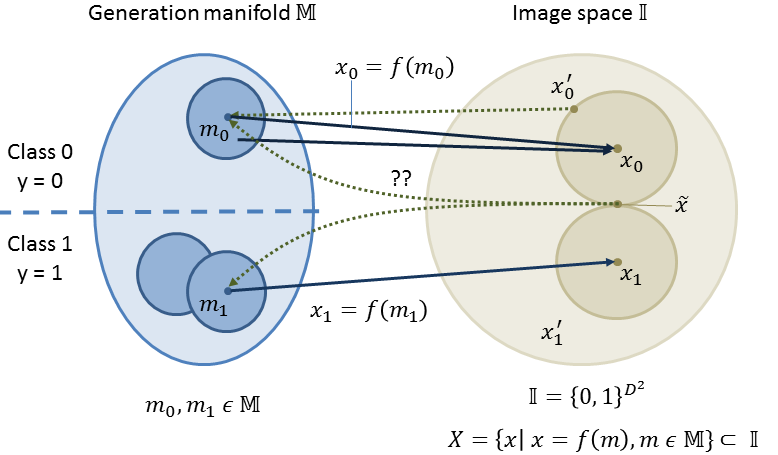}
	\caption{\label{fig:manifold-space}Conceptual summary of the image classification problem.}
\end{figure}
For a systematic analysis of adversarial examples, we need a closed-form representation of the decision boundary. 
Defining the decision boundary dependent on the generation manifold allows us to determine the distance of an image to the decision boundary in terms of the manifolds coordinates.
The generation function $f : \mathbb{M} \rightarrow \mathbb{I^*_f}$ maps from the generation manifold to the image space $\mathbb{I^*_f}$ restricted to images that can be generated by $f$.
Since all generations from a region around $m_i \in \mathbb{M}$ lead to identical images in $\mathbb{I}$, those cases are in-differentiable, i.e. $f$ is non-injective.
In addition, since not all images from $\mathbb{I}$ can be created by this function, some images $x \in \mathbb{I}$ do not have a corresponding coordinate $m$ in $\mathbb{M}$.
For images that cannot be directly generated through the generation function, but that are largely similar to images directly generated (e.g. one-pixel attacks $x'_i$ of image $x_i$), the association to the corresponding $m_i$ is implicit (dotted line).
Some images $\tilde{x}$ will have the property of being equally similar to two different $m$ even belonging to different classes.
For these ambiguous images, no association to a $m$ or even a class can be found.

\subsection{Dataset}
With our chosen generation manifold $\mathbb{M}$ as $(L, \alpha)$, our generation function maps to images of lines of varying length $L$ and rotation $\alpha$ (see \cref{fig:dataset}). 
Lines are centered on the image center, leading to a scenario where images always differ in at least 2 pixels because of symmetry.
A line-like structure is being used, to keep the generation manifold as simple as possible.
Finally, we define a simple decision boundary classifying images into 2 categories, where $y=1$ for $\alpha$ in the range from \ang{0} to \ang{40}, and $y = 0$ for all other cases.
The chosen range for $\alpha$ is arbitrary. 											

We generate 15 complete sets of all unique binary images $X = \mathbb{I^*_f}$ for 15 different generation functions $f$ by different manifold discretization (only allowing $\alpha$ in steps of \ang{0.5}, \ang{1.0} and \ang{2.0}) and image dimensions $D\times{D}$ ($16\times{16}$, $32\times{32}$, $48\times{48}$, $64\times{64}$ and $80\times{80}$). 
The length $L$ is varied between 12 Pixel and $D-2$, where $D$ is the width and the height of the image.
This procedure leads to varying numbers of unique images ($|X|$) as described in \cref{tab:numimages}.

\begin{table}[t]
	\begin{minipage}[]{0.65\textwidth}
		\centering
		\includegraphics[width=\textwidth]{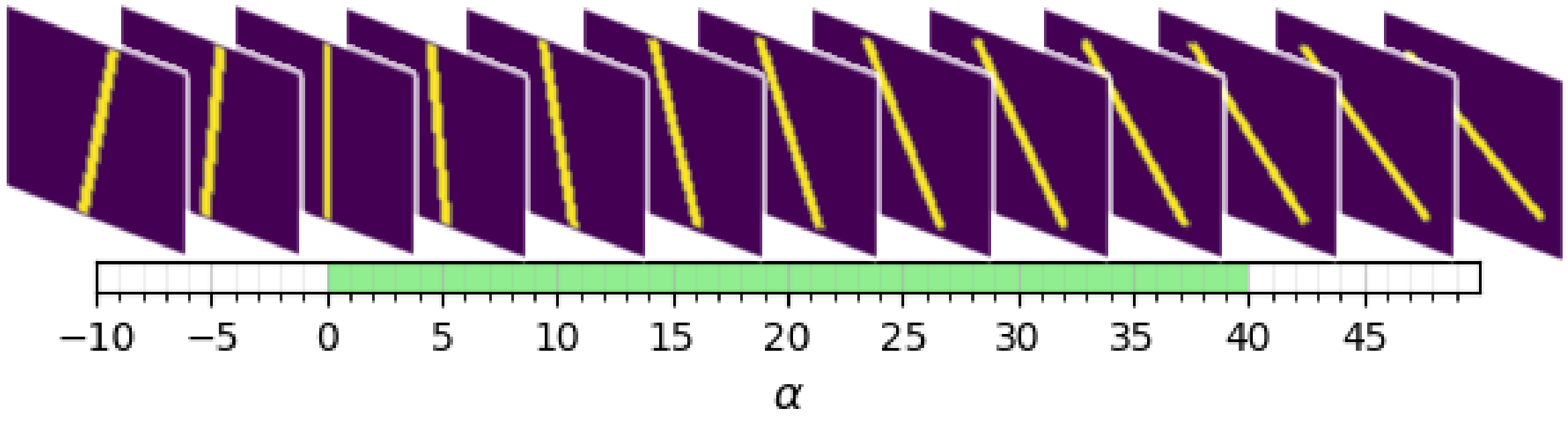}
		\captionof{figure}{\label{fig:dataset}Generated images from a subset. The bar below indicates the decision boundary. For the green range $y = 1$, $y = 0$ otherwise. Generated images are colorized for better contrast.}
	\end{minipage}\hfill
	\begin{minipage}[]{0.3\textwidth}
		\begin{tabular}{l|rrr}
			\toprule
			Dimensions & \ang{0.5}  & \ang{1.0}  & \ang{2.0} 
			\\
			\midrule
			$16\times{16}$ & 394 & 280 & 170 \\
			$32\times{32}$ & 2524 & 1394 & 716 \\
			$48\times{48}$ & 5244 & 2790 & 1436 \\
			$64\times{64}$ & 8116 & 4228 & 2158 \\
			$80\times{80}$ & 10998 & 5676 & 2880 \\
			\bottomrule
		\end{tabular}
		\caption{Number of unique images for different combinations of image dimensions and rotation stepsizes}
		\label{tab:numimages}
	\end{minipage}
\end{table}

\subsection{Training}
We train 5 models for each of the 15 synthetic combinations (\cref{tab:numimages}), resulting in a total of 75 trained networks.
Training repititions were performed to average out the stochastic properties of the training process.

Since we are looking at off-manifold images rather than the gereralization error of the networks, there is no need for a testing set.
Moreover, since we use every single image in a subset (i.e. every possible image) for training, our models are not restricted by the choice of training data.
All models feature the same architecture only differing in the dimensions of the input images.
We design a simple network architecture with three layers:
First, two 2D convolutional layers of size $3\times{3}$ with 32 channels and ReLU activation each followed by max pooling (stride 2) process the input.
On this, a fully connected layer with 128 units and ReLU activation is applied, followed by an output layer with 2 units and a Sigmoid activation.
We regulize by dropout ($p = 0.25$) just before the fully connected layer.

For optimization, we use the Adam optimizer with recommended parameters ($\beta_1 = 0.9$ and $\beta_2 = 0.999$) and a learning rate of 0.001.
Finally, binary crossentropy is used as the loss function.
We achieved an accuracy of 1.0 with all models indicating perfect convergence on the training dataset.

\begin{figure}[t]
	\centering
	\includegraphics[height=2cm]{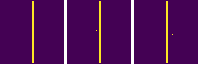}
	\caption{Left to right: image from dataset for \ang{0}; 2 confirmed exemplary adversarial candidates.}
	\label{fig:1pxattack}
\end{figure}
\subsection{Adversarial Data Creation}
The goal of creating adversarial data is to identify whether our trained networks can be fooled into predicting the wrong output $y$ for a given image $x$.
By flipping one pixel at a time, we perform an exhaustive search of all combinations of all images in $X$ (see \cref{fig:1pxattack} for examples). 
The total number of adversarial candidates $N_{adv} = N D^2$, where $N = |X|$ and $D$ being the width and length of the image.
Unlike Su \emph{et al.} \cite{Su.20180222}, we brute-force our way to a \emph{complete list of all possible adversarial examples} instead of finding single instances by optimization, which did not work for binary images.

\section{Results}
By testing the classification of all adversarial candidates, we found \emph{all networks} to be vulnerable to one-pixel-adversarial attacks.
All experiments were performed on 5 networks, so all values are averages over 5 networks.

\begin{figure}[h!]
	\centering
	\includegraphics[width=0.8\textwidth]{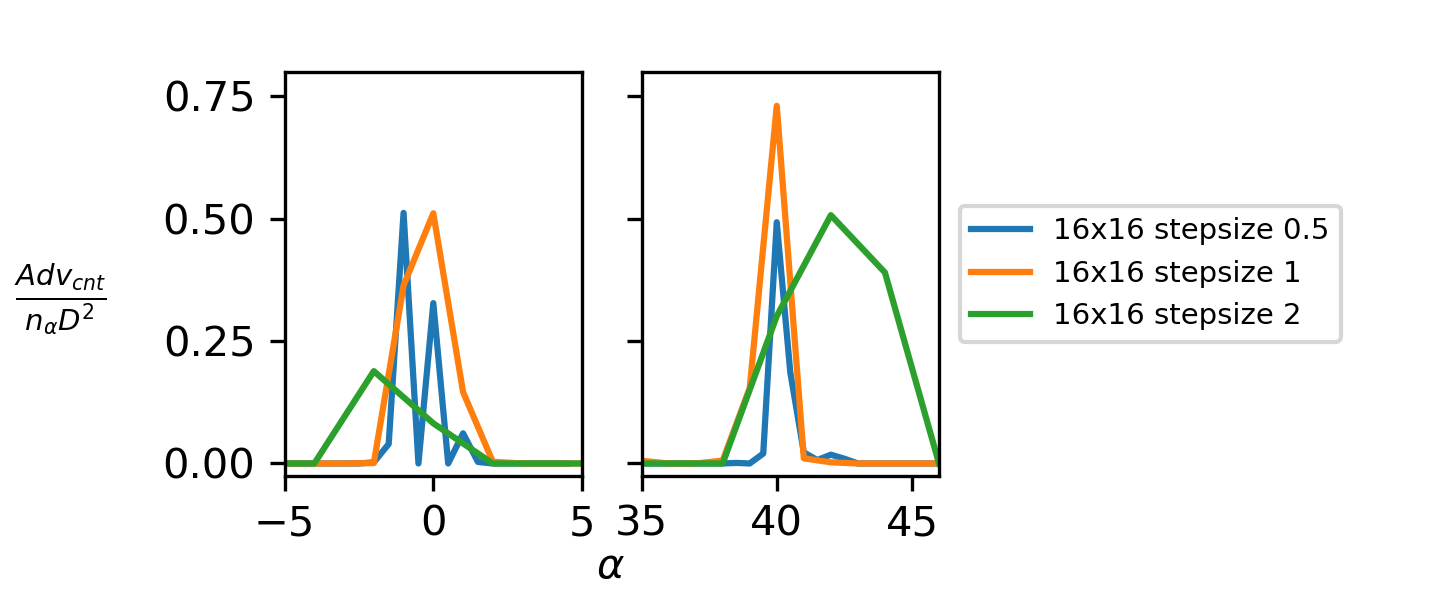}
	\includegraphics[width=0.8\textwidth]{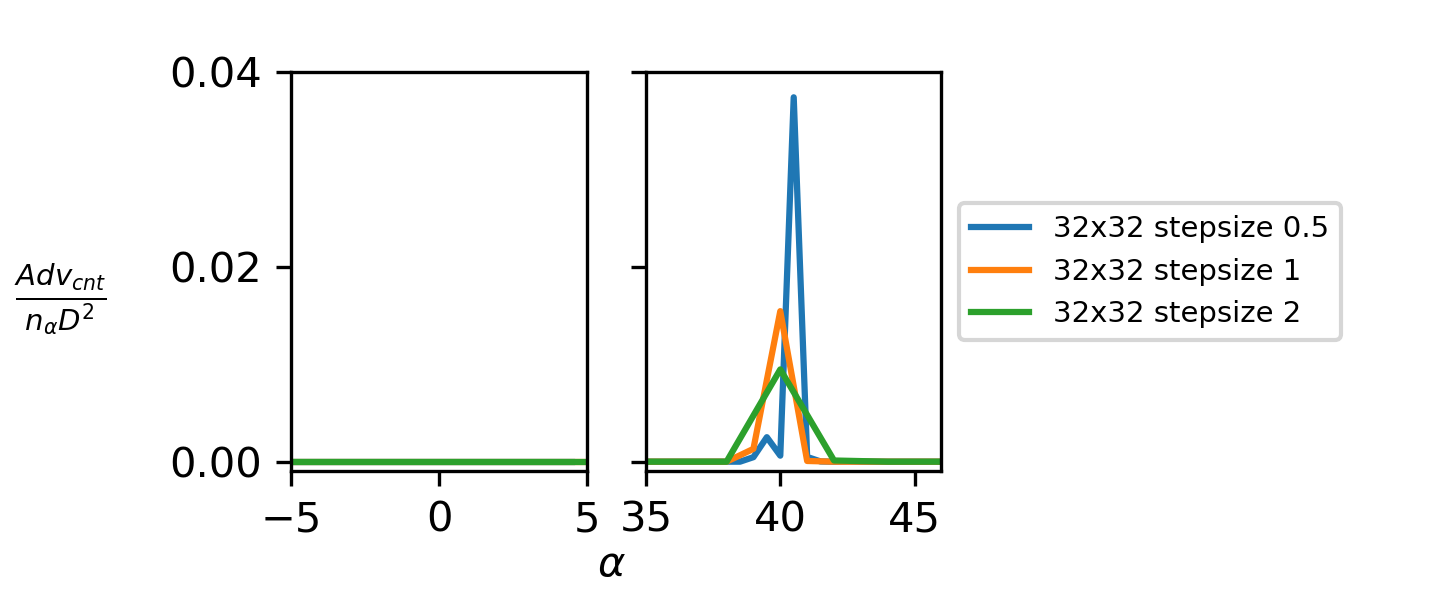}
	\includegraphics[width=0.8\textwidth]{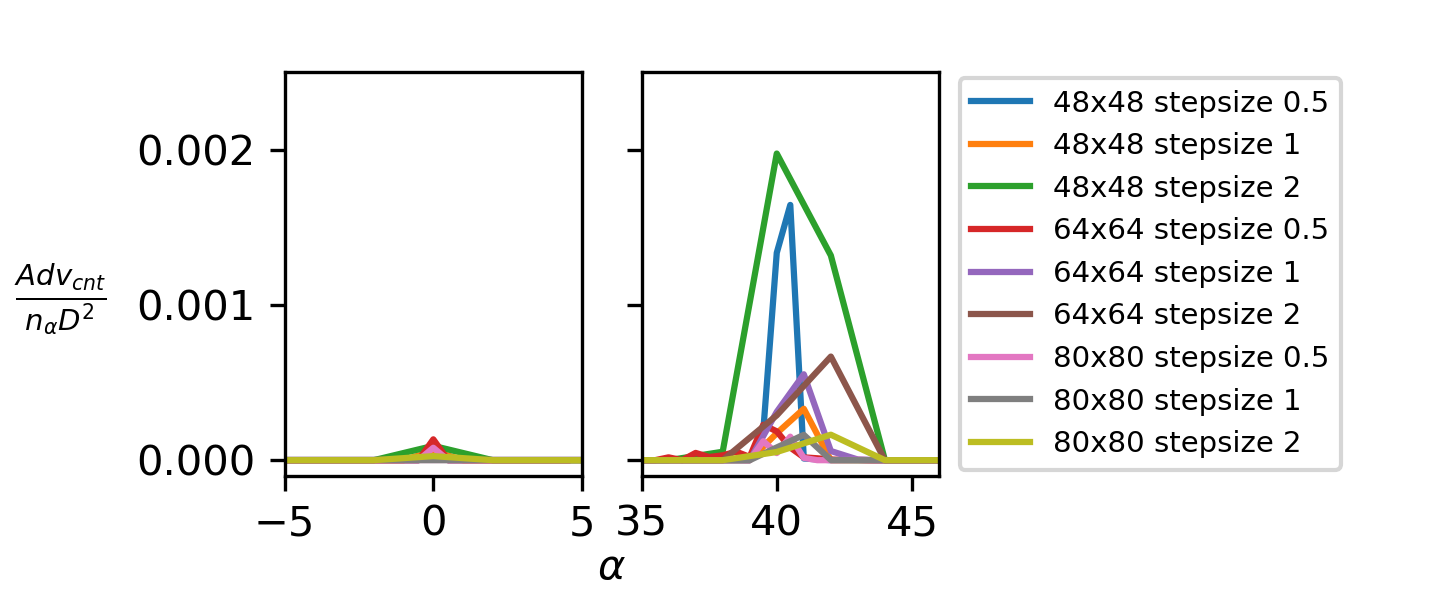}
	\caption{Comparison of average adversarial likelihood depending on the distance to decision boundary. Plots show the areas around $\alpha = \ang{0}$ and $\alpha = \ang{40}$. Note, in earlier versions (including the published version=, the y-axis was labeled incorrectly.}
	\label{fig:advcont-by-angle}
\end{figure}
\begin{figure}[th!]
	\centering
	\includegraphics[width=0.8\textwidth]{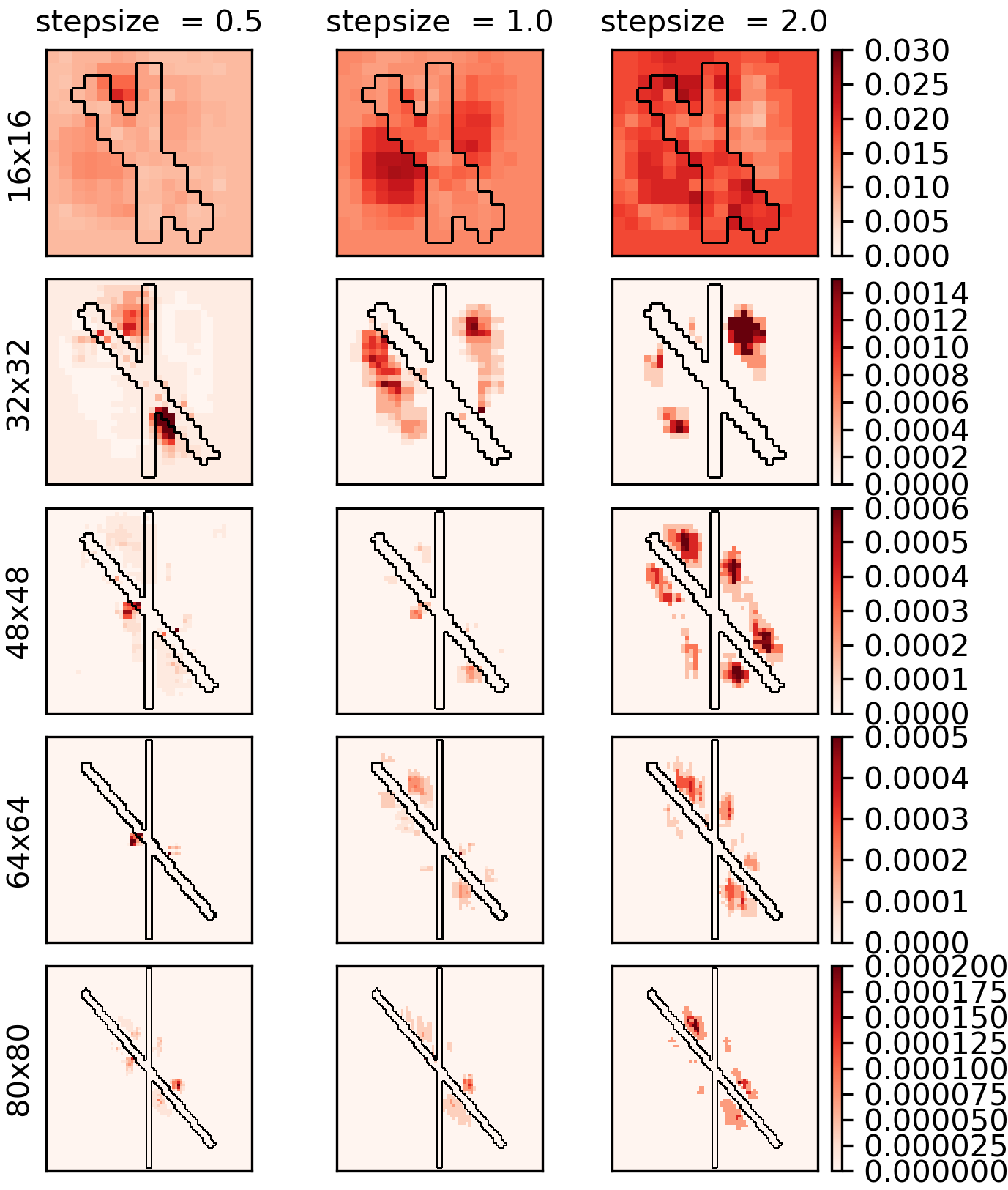}
	\caption{Heatmaps display the spatial likelihood for a flip to cause a misclassification, i.e. leading to an adversarial example; the ``X'' represent edges of two images on the decision boundary, i.e. $\alpha=\ang{0}$ and $\alpha=\ang{40}$; interestingly, high probabilities are often found in regions removed from the edges}
	\label{fig:heatmaps}
\end{figure}

We evaluated the relative number of adversarial examples w.r.t. the adversarial candidates.
We also determined this ratio $ADV_{cnt} / (N D^2)$ of actual adversarial examples to adversarial candidates dependent on the angle rotation $\alpha$ ($Adv_{cnt}/ (n_\alpha D^2)$, see \cref{fig:advcont-by-angle}) and on the pixel-position in the image (see \cref{fig:heatmaps}).
These ratios can also be interpreted as experimentally determined average likelihood of an image being adversarial given the specific conditions.

\begin{figure}[t]
	\centering
	\includegraphics[width=0.8\textwidth]{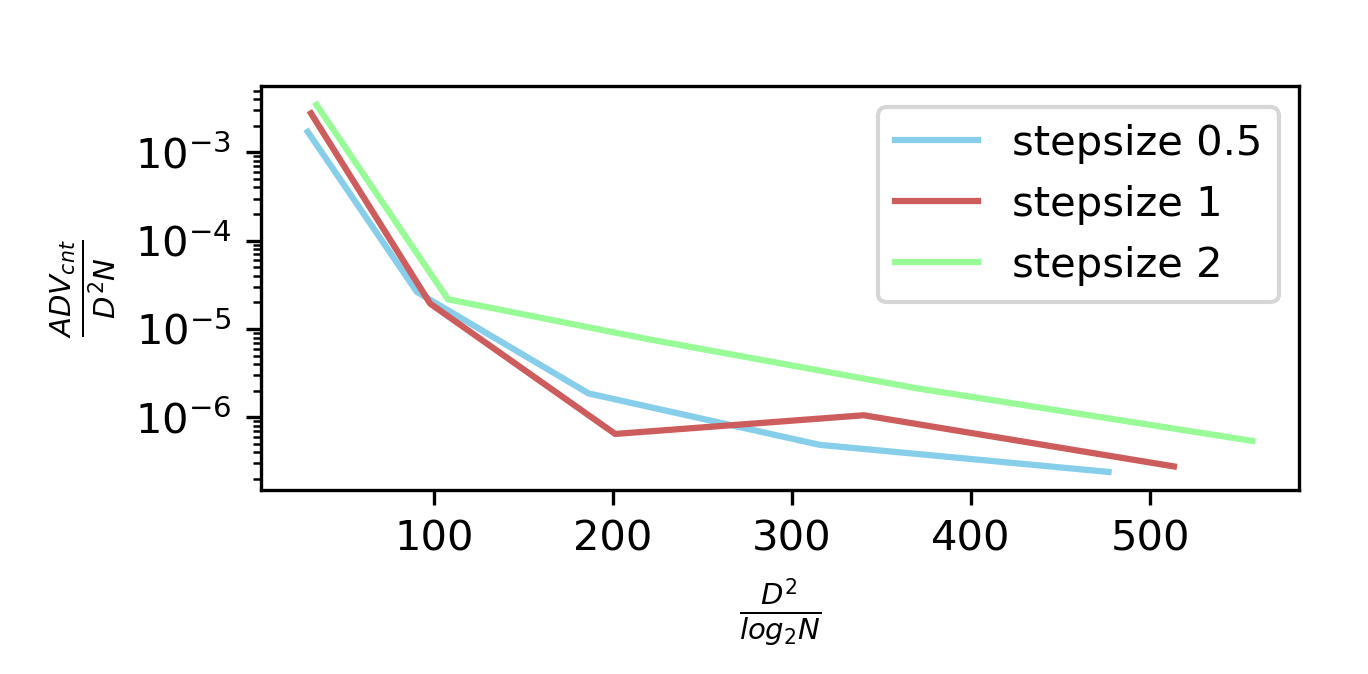}
	\caption{Information or Compression analysis}
	\label{fig:information-compression}
\end{figure}Figure \ref{fig:advcont-by-angle} shows the distribution of the relationship of misclassifications  at a particular angle.
Adversarial examples were only found around the decision boundary ($\alpha \approx \ang{0}$ or $\alpha \approx \ang{40}$).
With an increase of the image dimension $D$ the likelihood to find adversarial examples \emph{decreases}.

We created heatmaps, shown in figure \ref{fig:heatmaps}, to display the spatial likelihoods for a flip to cause a missclassification.
The ``X'' represented the edges of lines from two images, which belonged to the values of $\alpha$ on the decision boundary, i.e. $\alpha = \ang{0}$ and $\alpha = \ang{40}$.
Surprisingly, the highest likelihoods to cause an image to be confirmed as adversarial were not situated at positions, where the discrimination between classes got harder from an information perspective, but were removed from the edges, i.e. the overwhelming majority of images was \emph{not ambiguous}.
Increasing the image dimension led to more pronounced patterns and better overall robustness.
This can also be seen in \cref{fig:information-compression}.

Finally, we analyzed whether the relationship between the relative number of adversarial examples  and the ``possible redundancy'' of the reconstructable manifold information in image space exits.
The latter is the ratio of information contained in the reconstructable manifold and the image.
Higher ``redundancy'' seems to indicated a strong relationship to increased robustness to adversarial attacks.

Unlike other studies, these results were obtained for networks trained on \emph{``ALL'' possible images} that can be generated from the generation manifold and achieved an accuracy of $1.0$.

\section{Discussion and Conclusion}
This paper provides a systematic evaluation of one pixel adversarial attacks on convolutional neural networks.
By leveraging a simple toy CAI scenario against a simple yet perfect (accuracy $1.0$) convolutional neural network, we find one-pixel-adversarial-candidates with an astonishing  regularity.
These candidates are deliberately placed close to but off the manifold training images are drawn from.
In particular, we identify the vulnerable regions to be close to the decision boundary and not explainable by loss of informations caused by the introduction of the ``attacking pixel''.

In Future Work, we will generalize these observations to toy examples derived from other scenarios, increase the depth of the neural network and investigate causes for adversarial examples.

The systematic analysis of adversarial examples presented in this paper initiates a much needed process of understanding adversarial examples in medical images.

	\renewcommand{\bibsection}{\section*{References}} 
	\bibliographystyle{splncsnat}
	\begingroup
	\bibliography{bibliography-dk}
	\endgroup

\end{document}